\documentclass[letterpaper, 10 pt, conference]{ieeeconf}  

\IEEEoverridecommandlockouts                              

\usepackage{amsmath}
\usepackage{amsfonts}
\usepackage{graphicx}
\usepackage{booktabs}
\usepackage[ruled,vlined]{algorithm2e}
\usepackage[colorlinks=true, allcolors=blue]{hyperref}
\usepackage{subcaption}
\usepackage[font=small, tableposition=top]{caption}
\usepackage{siunitx}
\usepackage{refcount}
\newcommand\blfootnote[1]{%
  \begingroup
  \renewcommand\thefootnote{}\footnote{#1}%
  \addtocounter{footnote}{-1}%
  \endgroup
}

\usepackage{multirow}
\usepackage{censor}

\newif\ifblind
\blindfalse 

\ifblind
    \protect\RestartCensoring
\else
    \protect\StopCensoring
\fi

\title{\LARGE \bf
Navigating the Crowd: Non-linear MPC with Social Forces Dynamics for Human-Aware Robot Navigation
}

\author{
\censor{Stefano Trepella}$^{1}$, \censor{Andrea Ostuni}$^{1}$, \censor{Mauro Martini}$^{1}$, \censor{Pablo Pueyo}$^{2}$, \\ \censor{No\'e P\'erez-Higueras}$^{2}$, \censor{Marcello Chiaberge}$^{1}$, \censor{Fernando Caballero}$^{2}$, and \censor{Luis Merino}$^{2}$ 
\thanks{\xblackout{This work was partially supported by the SWIch action (P.R. F.E.S.R.2021/27 - D.G.R. n.19-6962) within the EMPATHY project, and by PoliTO Interdepartmental Centre for Service Robotics (PIC4SeR). It is also partially supported by the project AI-FUSE (SAIA202500X163851SV0) and COBUILD (PID2024-161069OB-C31), funded by the Spanish Research Agency and the Ministry of Science and the European Union (MCIN /AEI/10.13039/501100011033) and by ERDF "A way of making Europe"}} 
\thanks{\xblackout{$^{1}$Department of Electronics and Telecommunications, Politecnico di Torino, 10129, Torino, Italy.}
{\tt\footnotesize \xblackout{stefano.trepella@polito.it, andrea.ostuni@polito.it, mauro.martini@polito.it,  marcello.chiaberge@polito.it}}}%
\thanks{\xblackout{$^{2}$ School of Engineering, Pablo de Olavide University, Crta. Utrera km 1, Seville, Spain}
{\tt\footnotesize \xblackout{ppueyor@upo.es, noeperez@upo.es, fcaballero@upo.es, lmercab@upo.es}}%
}}

\begin{document}

\maketitle
\blfootnote{\copyright~2026 IEEE. Personal use of this material is permitted. Permission from IEEE must be obtained for all other uses, in any current or future media, including reprinting/republishing this material for advertising or promotional purposes, creating new collective works, for resale or redistribution to servers or lists, or reuse of any copyrighted component of this work in other works. This is the accepted version of a paper accepted for publication at the 2026 IEEE/RSJ International Conference on Intelligent Robots and Systems (IROS), September 27--October 1, 2026, Pittsburgh, PA, USA. The final published version will be available via IEEE Xplore.}
\thispagestyle{empty}
\pagestyle{empty}

\begin{abstract}
Safe and socially compliant navigation remains a fundamental challenge for autonomous robots operating in human-populated environments. Beyond collision avoidance, robots must anticipate human motion and respect personal space to ensure human comfort. 
Model Predictive Control (MPC) offers a robust alternative to classical and data-driven methods, although its effectiveness strongly depends on accurate human motion prediction and efficient computation.

This paper introduces SFM-NMPC, a Social Force Model-based Non-linear Model Predictive Control framework that embeds human motion prediction directly within the optimization loop. By incorporating the Social Force Model into the dynamic model of surrounding agents, the controller jointly predicts the trajectories of humans and robots over the prediction horizon, thereby enabling socially-aware planning. A tailored set of social cost functions guides the optimization toward human-compliant behaviors. Despite the increased model complexity, the proposed formulation runs in real time at \qty{20}{\Hz}.

Extensive simulated testing in crowded environments demonstrates that SFM-NMPC outperforms state-of-the-art baselines in social compliance metrics while maintaining efficient and smooth navigation. Visual trajectory analysis and an ablation study further highlight the contribution of the embedded SFM dynamics and social cost terms, confirming the effectiveness of the proposed approach for real-world social navigation.

\end{abstract}

\section{INTRODUCTION}
Autonomous robots operating in real-world environments must navigate both efficiently and safely, especially when interacting with humans. As the number of robots integrated into everyday tasks continues to increase, ranging from domestic robots~\cite{eirale2022marvin} to healthcare assistants in hospitals~\cite{holland2021service}, and autonomous wheelchairs for airport mobility~\cite{morales2017social}, the need for effective human-robot interaction models becomes critical~\cite{hussain2018autonomous}.

\begin{figure}[ht]
    \centering
    \includegraphics[width=\columnwidth, angle = 0]{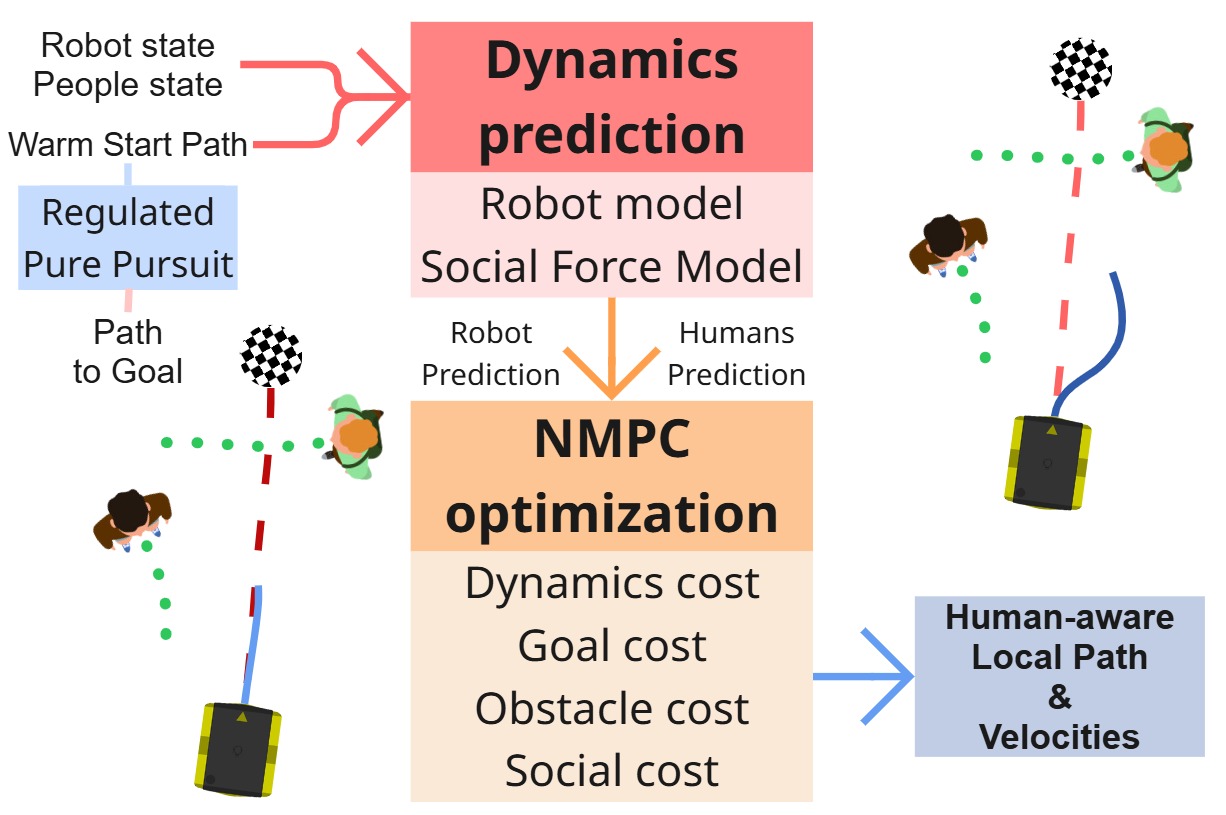}
    \caption{\footnotesize{SFM-NMPC embeds the Social Force Model in the prediction model of a Non-linear MPC controller as well as in the custom social cost terms to enhance human comfort and safety.}}
    \label{fig:first-page}
\end{figure}

Recent research on human-aware navigation has shown that socially compliant behavior improves human comfort and trust~\cite{guillen2023evolution}. However, notable challenges have been highlighted for real-world deployment in crowded spaces~\cite{singamaneni2024survey}.
In social contexts, robots must not only avoid physical obstacles, as in classic real-time control strategies ~\cite{DWA_paper}, but also maintain appropriate distances from other agents, predict their movements, and avoid abrupt or unpredictable maneuvers.
Emerging approaches such as Reinforcement Learning (RL)~\cite{SARL_paper} and Deep Learning (DL)~\cite{alahi2016social} have gained significant attention for their ability to learn predictive behaviors directly from data. However, these methods often require large datasets and substantial computational resources and may lack the transparency and robustness required in critical safety contexts.
A potential solution to overcome the limitation of these methods is Model Predictive Control (MPC) ~\cite{falcone2007predictive}, which optimizes the robot's trajectory over a finite time horizon by continuously updating decisions based on real-time observations of the robot's state and surroundings. A major advantage of MPC in social navigation is its ability to incorporate various constraints, including obstacle avoidance, velocity limits, and acceleration limits, within a unified optimization framework. As a result, MPC is particularly well-suited for real-time planning in human-populated environments.

However, to be truly effective, MPC must accurately predict human behavior in the prediction phase, which remains a key challenge in this area. Human movement is often unpredictable and influenced by a variety of social and psychological factors, making simple geometric models insufficient. As a result, people's trajectories are often precomputed even in already proposed social MPC formulations~\cite{everett2018motion}. 
The Social Force Model (SFM)~\cite{Helbing1995SocialFM}, originally developed to simulate crowd behavior, models human motion as the result of various social forces, such as attraction to goals, repulsion from other agents, and repulsion from obstacles. By incorporating SFM into robot navigation systems, robots can better understand and predict human motion, enabling them to adjust their trajectories in a socially aware manner.

In this work, we present the Social Force Model Non-linear MPC (SFM-NMPC), an efficient non-linear MPC that embeds the SFM in the dynamics model of people, predicting their trajectories along with the robot's within the optimization loop itself. The optimized commands are used to evolve the robot pose contextually and the humans' trajectories over the prediction horizon, yielding a more realistic forecast. The new implementation\footnote{\label{fn:repo}https://github.com/PIC4SeR/sfm-nmpc} allows the NMPC controller to be executed at \qty{20}{\Hz}, despite the complex dynamics of the crowd. Moreover, a custom set of social cost terms has been designed to promote optimization toward socially compliant commands, incorporating personal distance intrusion, relative headings, and an overall social work term derived from the SFM.
Figure \ref{fig:first-page} shows a visual summary of the proposed framework.
The extensive validation conducted across diverse crowded environments, evaluating classic navigation and social metrics, demonstrates the competitive advantage of the proposed method over many robust state-of-the-art baselines. 
Additionally, a visual comparison of trajectories is presented, along with an ablation study of the SFM-NMPC cost terms.

\section{RELATED WORK}
\label{sec:related_works}
Human trajectory prediction has been incorporated into robot navigation to improve human comfort and safety through Gaussian Processes~\cite{ellis2009modelling}, Inverse Reinforcement Learning~\cite{ziebart2009planning}, and neural network models such as RNNs and LSTMs~\cite{alahi2016social}. Despite their sophistication, such methods often require extensive data and computational resources to run in real time on the robot. ORCA~\cite{van2008reciprocal} is a popular obstacle avoidance algorithm adopted for human trajectory prediction~\cite{samavi2024sicnav}, although it treats people as standard moving obstacles. The Social Force Model (SFM)~\cite{Helbing1995SocialFM}, on the other hand, remains a widely used and interpretable alternative, inspired by physical crowd dynamics. Previous work shows that incorporating human trajectory models, such as SFM, improves navigation in dense crowds, balancing comfort, safety, and path fluency~\cite{bera2016glmp,rudenko2020human,gao2022evaluation}. Other studies have exploited the SFM in a reward function for Reinforcement Learning~\cite{martini2024adaptive}.
In contrast to the approaches mentioned above, Model Predictive Control (MPC)~\cite{mayne2000constrained} provides a robust framework to enforce robot dynamics, obstacle avoidance, and proxemic comfort.

Pedestrian-aware MPC solutions often use precomputed human trajectory forecasts~\cite{everett2018motion}. T-MPC++~\cite{degroot2024topologympc} samples homotopy classes to navigate complex spaces and select the lowest-cost, socially appropriate path. SHINE~\cite{SHINE_algo} improves social navigation by selecting topologically distinct guidance trajectories based on learned human preferences. Deep residual learning has also been integrated to improve efficiency using real crowd data~\cite{han2025dr}. In~\cite{fiasche2023towards}, the Social Force Model (SFM) is also embedded in an MPC for crowded scenes. However, these methods treat human trajectories as initial fixed predictions. Our approach integrates custom social cost terms with an efficient implementation of the SFM directly into the MPC loop, foreseeing contextually human and robot trajectories at high frequency at each step. This dynamics model supports real-time responsiveness to human behavior, aligning with social expectations while remaining interpretable and scalable.

\section{METHODOLOGY}
In this section, the proposed SFM-NMPC framework is formally presented. First, we define the adopted MPC formulation, the state and dynamics models used to propagate the prediction forward. Then, all the terms of the cost function are explained in detail. Finally, the overall optimization loop mechanism is described.

\subsection{Adopted MPC formulation}
\label{subsec:adopted_MPC_formulation}
We propose a local trajectory planner based on Non-linear Model Predictive Control (NMPC).
The original continuous-time system is considered as a discrete-time system using a direct method to ensure feasible optimization times. The control vector gives the variables to be optimized at each time step. Given the robot pose vector $\mathbf{p}_i^R \in SE(2)$, which contains the Cartesian coordinates and the yaw angle, and the robot control vector $\mathbf{u}_i \in \mathbb{R}^2$, which contains the linear and angular velocities, the transition function $g_t$ outputs the prediction of the pose vector in the next instant.
\begin{equation}
    \mathbf{p}_{i+1}^R = g_t( \mathbf{p}_i^R,\mathbf{u}_i)
\end{equation}
The transition function is a unicycle kinematic model that uses Euler forward propagation, 
\begin{equation}
    \mathbf{p}_{i+1}^R = \mathbf{p}_i^R + \mathbf{G}(\theta_i )\mathbf{u}_i\Delta t_s
\end{equation}
where $\Delta t_s$ is the prediction time step and $\mathbf{G(\theta_i )}$ is the control input mapping matrix. The main contribution of this work is that the robot pose $\mathbf{p}_i^R$, which is computed from the variables to be optimized, is also used to predict the agents' trajectories.
Considering a set of agents $\mathcal{A}$, the SFM computation is performed for every agent. Let $\mathbf{v}_i^{A_k}$, composed of $v_{x,i}^{A_k}$ and $v_{y,i}^{A_k}$ and $\mathbf{p}_i^{A_k}\in SE(2)$, composed of the agent Cartesian coordinates and the agent heading, define the velocity and pose of the $k$-th agent.
 For every prediction step $i$ and for every agent $k$, the agent's velocity $\mathbf{v}_{i}^{A_k}$ is updated by integrating the sum of acting forces,
\begin{equation}
    \mathbf{v}_{i+1}^{A_k} = \mathbf{v}_{i}^{A_k} + \left( \mathbf{F}_{soc}(\mathbf{p}_i^R, \mathbf{p}_i^{A_k}) + \mathbf{F}_{drive}(\mathbf{v}_i^{A_k}) \right) \Delta t_s 
\end{equation}
where $\mathbf{F}_{soc}$ represents the repulsive interaction forces and $\mathbf{F}_{drive}$ is a relaxation force representing the agent's resistance to deviate from its current state. 

\begin{figure}[t]
    \centering
    \includegraphics[width=.95\linewidth]{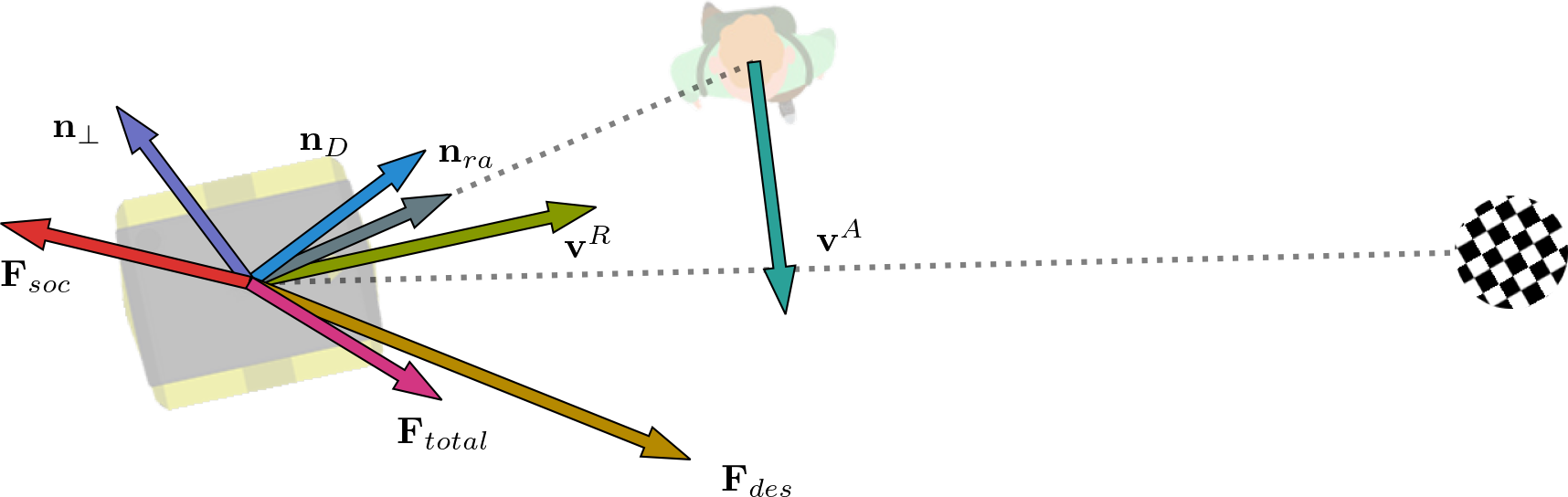}
    \caption{\footnotesize{The diagram of the social forces considered by the SFM\cite{Helbing1995SocialFM}.}}
    \label{fig:social_vector}
\end{figure}

Crucially, $\mathbf{F}_{soc}(\mathbf{p}_i^R, \mathbf{p}_i^{A_k})$ depends on the robot's pose $\mathbf{p}_i^R$  and the agent pose $\mathbf{p}_i^{A_k}$ at the current predicted time step (see Fig. \ref{fig:social_vector}). We utilize the following formulation  of the repulsive force to account for relative velocities: 
\begin{equation}
    \mathbf{F}_{soc} = A_{soc} ( f_v \mathbf{n}_D + f_{\theta} \mathbf{n}_{\perp} ) 
\end{equation}
where $\mathbf{n}_D$ is the unit interaction direction derived from the interaction vector $\mathbf{D} = \lambda (\mathbf{v}_i^R - \mathbf{v}_i^{A_k}) + \mathbf{n}_{ra}$, with $\mathbf{v}_i^R$ the linear velocity component of $\mathbf{u}_i$ which couples the relative velocity $\mathbf{v}$ and the unit direction $\mathbf{n}_{ra}$ between the entities poses $\mathbf{p}_i^R, \mathbf{p}_i^{A_k}$. $\mathbf{n}_{\perp}$ is the perpendicular vector to $\mathbf{n}_D$.  The scalar magnitudes $f_v$ and $f_{\theta}$ represent longitudinal braking and lateral steering forces. These are exponential decay functions of $\|\mathbf{r}_{ra}\|$, the Euclidean distance between the robot and agent $k$, and the angular displacement $\theta$ between the relative position and interaction directions. 

The agent's position is computed after the velocity update using Semi-Implicit Euler integration:
\begin{equation}
    \mathbf{p}_{i+1}^{A_k} = \mathbf{p}_{i}^{A_k} + \mathbf{v}_{i+1}^{A_k} \Delta t_s
\end{equation}
noting that the heading is given by the $\mathbf{v}_{i+1}^{A_k}$ direction.

\subsection{Cost function}
\label{subsec:cost_function}
Forward propagation optimizes only the velocity commands.
The velocity command generation is formulated as a non-linear least-squares (NLS) optimization problem, using Ceres \cite{Agarwal_Ceres_Solver_2022} as the solver.
If we use $\chi$ to denote the set of velocity vectors $\mathbf{u}_i$, the main objective of the optimization loop is to obtain the optimal sequence of velocity commands $\chi^*$: 
\begin{equation}
    \begin{aligned}    
         \chi^* = \min_{\chi}  J(\chi)  := J_{\mathrm{obs}} + J_{\mathrm{goal}} + J_{\mathrm{dyn}} + J_{\mathrm{soc}} \\
        \begin{alignedat}{3}
        {\scriptstyle\text{s.t.}} \quad
          & {\scriptstyle \mathbf{p}_{i+1}^R = \mathbf{p}_i^R + \mathbf{G}(\theta_i )\mathbf{u}_i\Delta t_s} & {\scriptstyle \forall i = 0, \ldots, N_p} \\
          & {\scriptstyle\mathbf{p}_{i+1}^{A_k} = \mathbf{p}_{i}^{A_k} + \mathbf{v}_{i+1}^{A_k} \Delta t_s} & {\scriptstyle \forall k \in \mathcal{A},  \forall i = 0, \ldots, N_p}\\
          & {\scriptstyle\mathbf{v}_{i+1}^{A_k} = \mathbf{v}_{i}^{A_k} + \left( \mathbf{F}_{soc} + \mathbf{F}_{drive} \right) \Delta t_s} \quad &  {\scriptstyle\forall k \in \mathcal{A},  \forall i = 0, \ldots, N_p-1} \\
          & {\scriptstyle v_{\text{lb}} \le v_i \le v_{\text{ub}}} & {\scriptstyle \forall i = 0, \ldots, N_p-1} \\
         & {\scriptstyle \omega_{\text{lb}} \le \omega_i \le \omega_{\text{ub}}} & {\scriptstyle \forall i = 0, \ldots, N_p-1} \\
        \end{alignedat}
    \end{aligned}
    \label{eq:cost_function}
\end{equation}
$N_p$ represents the number of points in the prediction horizon, and the robot velocities to be optimized are bounded between $\left[v_{lb},v_{ub}\right]$ for the linear velocities and $\left[\omega_{lb},\omega_{ub}\right]$ for the angular ones.

$J(\chi)$ in Eq. \ref{eq:cost_function} represents the sum of different costs to be considered; these represent the competing objectives for the robot to achieve; each term is weighted by a different factor $\alpha$, aiming to balance obstacle avoidance, social behavior, and smooth trajectories.
The specific weight values used for each term in the cost function are available in our public code repository\textsuperscript{\ref{fn:repo}}.
$J_{\mathrm{obs}}$ represents the cost given by the local costmap $\mathcal{M}_{local}$ of dimension \qtyproduct{6 x 6}{\metre} during optimization, aiming to increase the distance from high-cost zones,
\begin{equation}    
J_{\mathrm{obs}} = \alpha_{\mathrm{obs}}\sum_{i=1}^{N_p} \mathcal{C}_{map}(\mathcal{M}_{local},\mathbf{p}_i^R) 
\end{equation}
where $\mathcal{C}_{map}(\mathcal{M}_{local},\mathbf{p}_i)$ is given by the Ceres bi-cubic interpolation, using the 16 surrounding grid cells (with a resolution of \qty{0.05}{\metre}) compared to the i-th pose $\mathbf{p}_i$ to compute a smooth cost value.    

$J_{\mathrm{goal}}$ encompasses four costs given to push the robot towards the local goal, dynamically obtained from global path, as better explained in Section \ref{subsec:optimization_loop}: $J_{\mathrm{p\_follow}}$ pushes the robot close to the last valid path point, $J_{\mathrm{p\_align}}$ aims at aligning the predicted points with the given path, $J_{g\_align}$ tends to rotate the robot to face the goal, and finally $J_{\mathrm{g\_reach }}$ is used to attract the robot toward the final goal strongly.

$J_{\mathrm{dyn}}$ includes two costs aimed at making the robot trajectory smoother and less jittery;
$J_{\mathrm{vel}}$ encourages the robot's linear velocity to be close to a desired value while $J_{\mathrm{feas}}$ limits aggressive acceleration by penalizing the difference between consecutive velocity commands.

$J_{\mathrm{soc}}$ is composed of different social penalties to minimize discomfort on the agents, which are described below.

\subsubsection{Proxemics cost}
$J_{\mathrm{\rho}}$ is a purely distance-based repulsion. It penalizes the robot for approaching an agent too closely. It is enforced as a sum of exponential decay functions over all agents visible at any time, acting only on the Euclidean distances between their positions,
\begin{equation}    
J_{\mathrm{\rho}} = \alpha_{\mathrm{\rho}}\sum_{i=1}^{N_p}\sum_{k=1}^\mathcal{N_A} \beta \cdot\exp\left(-\frac{\lVert \mathbf{p}_{xy,i}^R - \mathbf{p}_{xy,i}^{A_k} \rVert^2}{d_0^2}\right)
\end{equation}
where $d_0$ is the characteristic proxemic distance and $\beta$ is a scaling factor.

\subsubsection{Social Work}
$J_{\mathrm{work}}$ minimizes the social force exerted by the robot on agents and vice versa
\begin{equation}
J_{\mathrm{work}} = \alpha_{\mathrm{work}} \sum_{i=1}^{N_p} \left( \sum_{k=1}^\mathcal{N_A} ||\mathbf{F}_{robot \to agent_k}||^2 \right)
\end{equation}
where $\mathcal{N_A}$ is the number of involved agents.
$F_{robot \to agent_k}$ is the sum of two forces: 
a braking force $f_{v}$ in the longitudinal direction, and a steering force $f_{\theta}$ in the lateral direction, according to the Helbing formulation of the SFM \cite{Helbing1995SocialFM}.

\subsubsection{Heading social cost}
a heading cost term $J_{\mathrm{s\theta}}=J_{\mathrm{s\theta long}}+J_{\mathrm{s\theta cross}} $ to boost predictive and socially compliant planning of the robot. $J_{\mathrm{s\theta long}}$ penalizes the robot for heading towards or in the same direction as the nearest agent,  steering the robot away. It is active for the closest agent ($k^*$) at the start of the prediction.
\begin{equation}
\begin{aligned}    
    J_{\mathrm{s\theta long}} = \alpha_{\mathrm{s\theta}} \sum_{i=1}^{N_p}  e^{-d^2/d_s^2} \cdot \Big(\underbrace{\operatorname{sp}\!\big(\cos(\theta_i - \phi_{k^*})\big)}_{\text{position alignment}} +\\
     \alpha_v \cdot \underbrace{\operatorname{sp}\!\big(\cos(\theta_i - \psi_{k^*})\big)}_{\text{velocity alignment}}\Big)
\end{aligned}
\end{equation}
where $\phi_{k^*}$ is the bearing to agent $k^*$ and $\psi_{k^*}$ is the agent's heading. The $\operatorname{sp}$ denotes the Softplus smoothing function, which acts as a smooth ReLU, activating the penalty only when the robot points toward the agent or travels in the same direction.
 
$J_{\mathrm{s\theta cross}}$ addresses specifically perpendicular crossing scenarios using two different components,
\begin{equation}
\begin{aligned}   
  J_{\mathrm{s\theta cross}} = \alpha_{\text{cross}} \sum_{i=1}^{N_p} e^{-d^2/d_s^2} \Big(\underbrace{v_i \cdot \sin^2(\theta_i - \psi_{k^*})}_{\text{speed penalty}} + \\ \alpha_{\text{bear}} \cdot \underbrace{\operatorname{sp}\!\big(\mathrm{c} \cdot \omega_i \cdot s\big) \cdot \sin^2(\theta_i - \psi_{k^*})}_{\text{steering penalty}}\Big),
\end{aligned}
\end{equation}
where $\sin^2(\Delta\theta)$ gates the cost to peak at 90$^\circ$ crossings and vanish for the same heading. The scalar $\mathrm{c} = \mathbf{n}_{ra} \times  \mathbf{n}_\psi$  is the 2D cross product of the robot-to-agent vector with the agent's heading, indicating whether the agent approaches from the left or right. The multiplier $\operatorname{sp}\!\big(\mathbf{c} \cdot \omega_i \cdot s\big)$ penalizes angular velocity in the wrong rotational direction (i.e., turning in front of rather than behind the crossing agent), where $s$ is a scaling factor. 

\begin{algorithm}[t]
\caption{SFM Non-linear MPC}
\SetAlgoLined
\DontPrintSemicolon
\SetKwInOut{Input}{Input}
\SetKwInOut{Output}{Output}

\Input{$\mathcal{P}_{global}$, $\mathcal{M}_{local}$, $\mathcal{A}$, $p_0$, $\mathcal{T}_{prev}$, $\chi_{prev}$}
\Output{$\left\{  \chi^*, \mathcal{T}^* \right\}$}

Compute reference trajectories with $(\mathcal{T}_{ref}, U_{ref}) = RPP(P_{global}, L_d, p_0)$\;

Warm-start control sequence $\chi_0 = (U_{ref},\chi_{prev})$\;
Warm-start trajectory $\mathcal{T}_0 = (\mathcal{T}_{ref},\mathcal{T}_{prev})$

Build the cost function in Eq.\ref{eq:cost_function} and propagate the robot pose and the agents forward using the SFM.

Solve the non-linear least-squares problem with Ceres starting from $\chi_0$ and $\mathcal{T}_0$ to obtain $\chi^*$\;
Roll out $\mathcal{T}^*$ by forward-propagating $p_0$ with $\chi^*$\;
\Return{$(\chi^*,\mathcal{T}^*)$}\;
\label{algo:control_loop}
\end{algorithm}

\subsection{Optimization loop}
\label{subsec:optimization_loop}
The main control loop is described in Algorithm
\ref{algo:control_loop}. 
Let $\mathcal{P}_{global}$ denote the global path generated at the start of navigation. We used a Regulated Pure Pursuit algorithm to extract a local reference trajectory, $\mathcal{T}_{ref}$. The Regulated Pure Pursuit algorithm acts on a set of poses obtained by pruning $\mathcal{P}_{global}$ using the robot's current pose, $\mathbf{p}_0$, and a dynamic lookahead distance $L_d$. The pruned poses allow $\mathcal{T}_{ref}$ to provide a set of robot-reachable poses and a series of initial control inputs $U_{ref}$.
The optimizer evaluates the overall cost function at each iteration, using information provided by the navigation stack.
If a previous solution $\chi_{prev}$ is available (i.e. after any successful optimization), $\chi_0$ is determined by combining $U_{ref}$ and $\chi_{prev}$ using $\alpha_{com}$ to weigh the reference command against the previous optimization result. The same operation is done if a previous trajectory $\mathcal{T}_{prev}$ exists, obtaining a $\mathcal{T}_0$ by $\mathcal{T}_{ref}$ and $\mathcal{T}_{prev}$.
After computing initial estimates of velocities and poses, the optimization is performed at \qty{20}{\Hz}, following the receding-horizon principle.
The number of prediction points $N_p$ is determined by the desired temporal prediction horizon $T_p$ and $\Delta t_s$, with the ratio $N_p=T_p/\Delta t_s$.
Two additional variables are introduced to limit the computational cost of the algorithm. The first variable is the control horizon $T_c$, the time interval during which the velocity command can be modified within the prediction horizon, after $T_c$ the last command is kept constant until the end of the prediction horizon. The number of control points is then given by the ratio $N_c=T_c/\Delta t_s$. The second variable is the parameter block length $B$, which assumes that the variable to be optimized is kept constant for $B$ time steps.
At each time step, the optimizer forward-propagates the robot's pose using the current velocity $\mathbf{u}_i$. It also propagates each agent's pose and computes the cost function for the i-th step. The individual cost terms obtained are then summed and solved simultaneously using Ceres at the final prediction point, with box constraints to ensure that velocities remain feasible throughout the optimization. The optimized velocity vector $\chi^*$ is used to generate the corresponding optimized trajectory $\mathcal{T}^*$. The control input applied to the robot is the first element of $\chi^*$, while the remaining elements initialize the velocity variables for the next iteration.

\section{EVALUATION METHODOLOGY}
\begin{figure}[!ht]
    \centering
    \subfloat[Open Space World - Open space scenarios.]{
       \includegraphics[width=0.82\columnwidth]{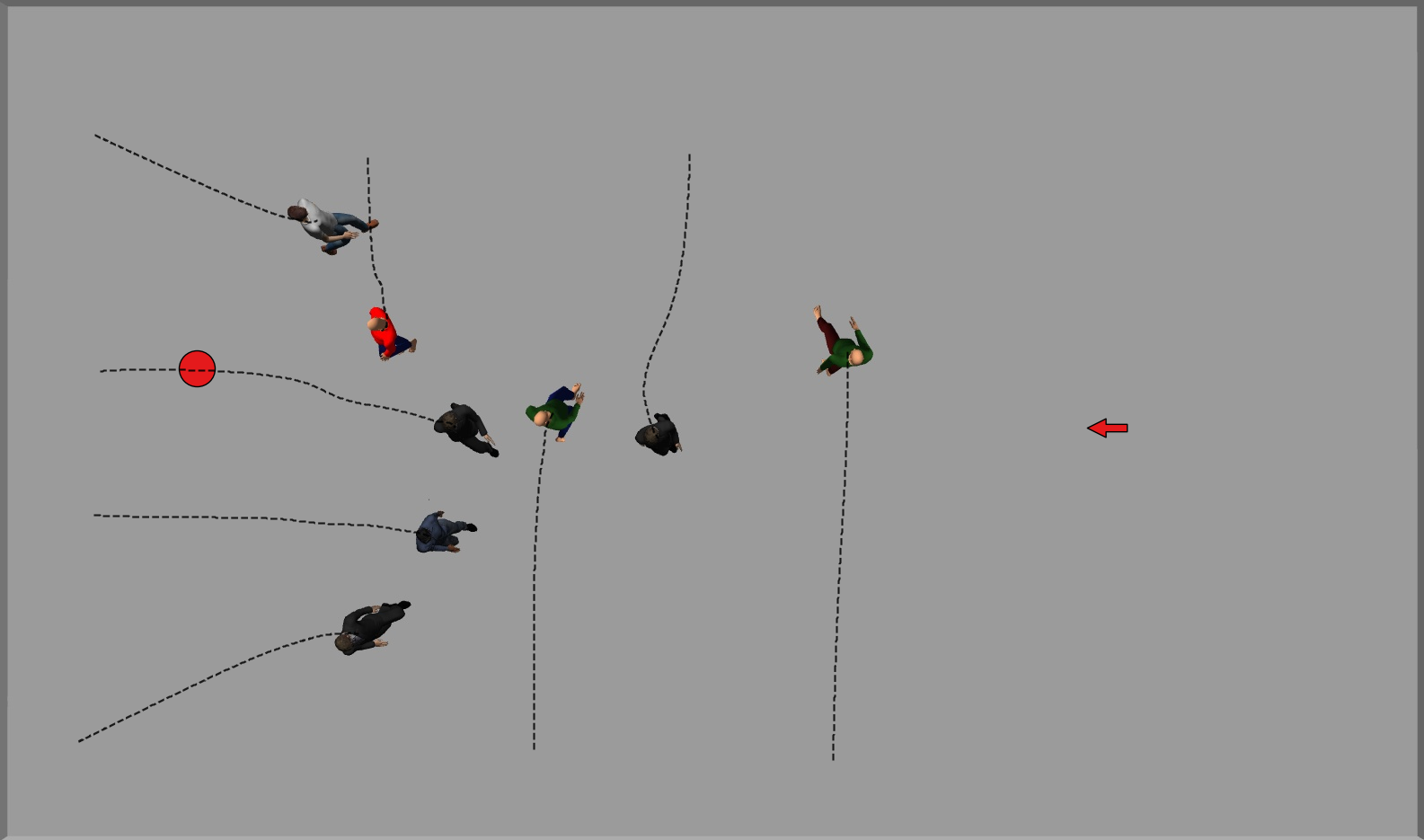}
       \label{fig:crowded_env}
    } \\
    \subfloat[Mixed World - Corridors, small rooms, and mixed scenarios.]{
       \includegraphics[width=0.82\columnwidth]{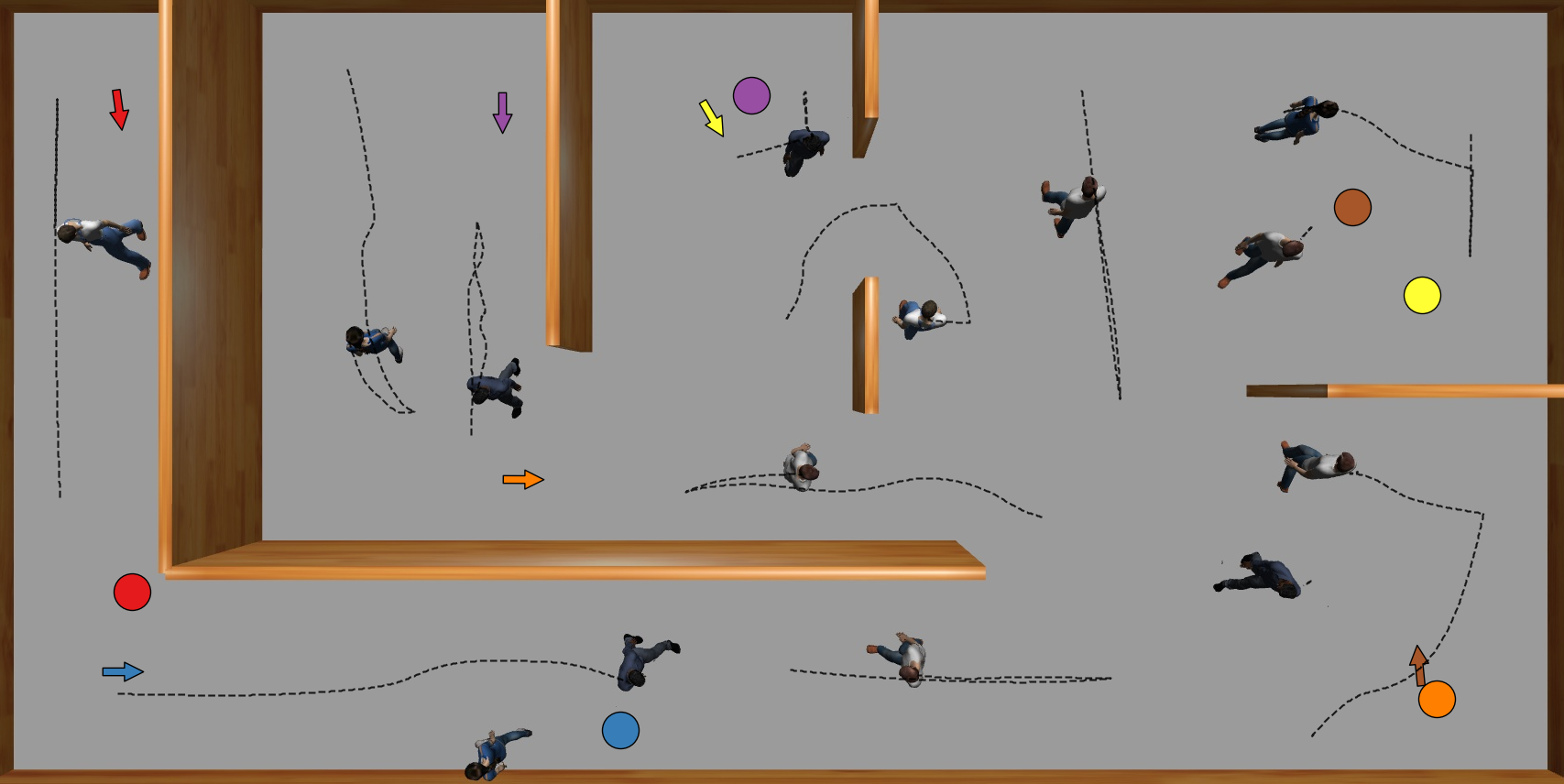}
       \label{fig:social_nav}
    }
 \caption{\footnotesize{Maps used for the tests and trajectories of robots and humans across the different scenarios.}}
    \label{fig:scenario_maps}
\end{figure}

We evaluated the proposed SFM-NMPC local planner and a selection of popular state-of-the-art baselines across a set of social scenarios, encompassing all relevant situations reported in the literature, from open, crowded spaces to constrained corridors and passages. 
For human motion, we used the HuNavSim \cite{perez2023hunavsim} simulation and benchmarking suite in Gazebo to enable realistic physical interactions with agents and obstacles and to collect relevant social navigation metrics.
Two separate environments shown in Figure \ref{fig:scenario_maps} have been designed to test all the algorithms in diverse conditions. 

Figure \ref{fig:crowded_env} represents the first environment, designed to be open and to focus more on behavior in response to agents under ideal conditions. Figure \ref{fig:social_nav} presents the second environment, a mix of corridors, rooms, and narrow passages to create more realistic and challenging scenarios for the robot.
In the first open-space map, we define three distinct scenarios. These scenarios are a crossing and a passing interaction with four pedestrians, and a generally crowded navigation with eight people mixing the two conditions. The crowded scenario is the one shown in Figure \ref{fig:crowded_env}.
In the second map, six structured navigation episodes have been designed, including human passing and overtaking in corridors, traversing crowded rooms, navigating narrow passages, and turns typically present in real indoor environments.
Each algorithm was tested 30 times per scenario to ensure repeatability and consistent results.
The experiments were performed on a workstation equipped with a Intel i9-12900K CPU and 64 GB of RAM.
\subsection{Baselines}
A wide set of relevant state-of-the-art algorithms for human-aware local planning has been selected as competitive baselines to compare and validate the proposed SFM-NMPC method:
\begin{itemize}
    \item \textbf{$\mathrm{MPPI_{GSC}}$}: a Model Path Predictive integral (MPPI)~\cite{MPPI_paper} implementation from  Nav2~\cite{macenski2020marathon} with a Gaussian social costmap plugin~\cite{social_navigation_layers};
    
    \item \textbf{$\mathrm{DWB_{GSC}}$}: a Dynamic Window Approach~\cite{DWA_paper} improved implementation from the Nav2~\cite{macenski2020marathon} library with a Gaussian social costmap plugin~\cite{social_navigation_layers};
    
    \item ORCA: Optimal Reciprocal Collision Avoidance~\cite{van2008reciprocal}, leveraging the RVO2 library;
    
    \item SARL: Socially Attentive Reinforcement Algorithm~\cite{SARL_paper};
    
    \item Pure SFM: the Social Force Model, using the computed forces directly to determine the next velocity for the robot~\cite{Helbing1995SocialFM};
    
    \item NMPC: the base implementation of our NMPC without SFM dynamics and any social cost.
\end{itemize}
They were selected from recent, widely used algorithms with available and reliable implementations, enabling a comprehensive, competitive, and realistic comparison. We implemented and adapted all algorithms as Nav2~\cite{macenski2020marathon} controller plugins to ensure consistency and a fair comparison, integrating them with a global A* path planner and costmaps. The global path was generated only once at the beginning of the episode, without replanning. For SARL, ORCA, and Pure SFM, waypoints are dynamically generated along the global path with a lookahead distance of 1\si{m} to enable traversal of the entire map.
To enable MPPI and DWA to account for social behavior, a Gaussian social costmap plugin~\cite{social_navigation_layers} that incorporates the person's pose and velocity has been employed. 
The running frequency of the local controllers has been set to \qty{20}{\Hz}, matching the SFM-NMPC operating frequency, and all common Nav2 parameters have been kept constant across the different algorithms. For the NMPC, the prediction horizon and time step have been set to $T_p = 2s $ and $\Delta t_s = 0.1s$.

\subsection{Evaluation Metrics}
All algorithms have been evaluated using a set of quantitative metrics that assess both navigation efficiency and social compliance. This set includes three metrics that assess traditional navigation performance: the success rate (SR), the time to reach the goal (TTG) and the path length (PL). Besides, to evaluate the local planners social behavior, two additional metrics have been considered: the social work per step ($SW_{step}$) and the average minimum distance to the closest person (AMD).
The $SW_{step}$ has been evaluated as the average of the social forces generated by the robot along its trajectory, according to the formulation of the SFM~\cite{Helbing1995SocialFM}.  
The success rate (SR) is calculated across all experimental trials.

\section{RESULTS}
In this section, the experimental results are presented and discussed. First, we analyze the quantitative metrics and the proxemics results. Then, a visual comparison of the trajectories is performed. Finally, an ablation study is conducted to investigate the contribution of each social cost term to the overall result. The video in the multimedia attachment provides a clearer visualization of the results.

\begin{table}[t]
	\centering
	\caption{\footnotesize{Summary of social navigation metrics for each navigator in the Mixed and Open Space scenarios, as well as overall. 
    }}
	\label{tab:algorithm_metrics}
	\resizebox{\columnwidth}{!}{%
	\begin{tabular}{llccccc}
    	\toprule
    	\textbf{Map} & \textbf{Method} & \textbf{{SR (\%)}} $\mathbf{{\uparrow}}$ & \textbf{{PL (m)}} $\mathbf{{\downarrow}}$ & \textbf{{TTG (s)}} $\mathbf{{\downarrow}}$ & $\mathbf{{SW_{{step}}}}$ $\mathbf{{\downarrow}}$ & \textbf{{AMD (m)}} $\mathbf{{\uparrow}}$ \\
    	\midrule
    	\multirow[t]{7}{*}{\textbf{Open}} & \textbf{$\mathrm{DWB_{GSC}}$} & $\mathbf{100.00}$ & $\mathbf{10.36}$ & $\mathbf{15.83}$ & $9.96$ & $1.60$ \\
    	\textbf{Space} & \textbf{$\mathrm{MPPI_{GSC}}$} & $\underline{99.00}$ & $\underline{10.56}$ & $17.57$ & $9.83$ & $1.52$ \\
    	\textbf{} & \textbf{$\mathrm{ORCA}$} & $81.00$ & $11.71$ & $26.39$ & $9.10$ & $1.63$ \\
    	\textbf{} & \textbf{$\mathrm{Pure\ SFM}$} & $\mathbf{100.00}$ & $11.00$ & $17.47$ & $9.73$ & $\underline{1.71}$ \\
    	\textbf{} & \textbf{$\mathrm{SARL}$} & $96.00$ & $15.94$ & $29.47$ & $\mathbf{8.04}$ & $1.70$ \\
    	\textbf{} & \textbf{$\mathrm{NMPC}$} & $\mathbf{100.00}$ & $10.88$ & $\underline{16.73}$ & $10.27$ & $1.53$ \\
    	\textbf{} & \textbf{SFM-NMPC} & $\mathbf{100.00}$ & $11.13$ & $18.17$ & $\underline{8.17}$ & $\mathbf{1.73}$ \\
    	\midrule
        \multirow[t]{7}{*}{\textbf{Mixed}} & \textbf{$\mathrm{DWB_{GSC}}$} & $42.00$ & $9.04$ & $14.27$ & $16.02$ & $0.99$ \\
    	\textbf{} & \textbf{$\mathrm{MPPI_{GSC}}$} & $\underline{91.00}$ & $9.28$ & $16.56$ & $13.48$ & $\underline{1.00}$ \\
    	\textbf{} & \textbf{$\mathrm{ORCA}$} & $16.00$ & $\mathbf{5.99}$ & $\mathbf{10.45}$ & $19.90$ & $0.98$ \\
    	\textbf{} & \textbf{$\mathrm{Pure\ SFM}$} & $31.00$ & $17.34$ & $31.43$ & $13.80$ & $\mathbf{1.02}$ \\
    	\textbf{} & \textbf{$\mathrm{SARL}$} & $46.00$ & $\underline{7.98}$ & $\underline{11.91}$ & $17.19$ & $0.93$ \\
    	\textbf{} & \textbf{$\mathrm{NMPC}$} & $81.00$ & $9.26$ & $16.50$ & $\underline{13.27}$ & $0.88$ \\
    	\textbf{} & \textbf{SFM-NMPC} & $\mathbf{97.00}$ & $9.96$ & $18.70$ & $\mathbf{11.96}$ & $0.99$ \\
    	\midrule
    	\multirow[t]{7}{*}{\textbf{Overall}} & \textbf{$\mathrm{DWB_{GSC}}$} & $70.00$ & $\mathbf{9.54}$ & $\mathbf{14.86}$ & $14.00$ & $1.19$ \\
    	\textbf{} & \textbf{$\mathrm{MPPI_{GSC}}$} & $\underline{94.00}$ & $\underline{9.71}$ & $16.90$ & $\underline{12.26}$ & $1.17$ \\
    	\textbf{} & \textbf{$\mathrm{ORCA}$} & $38.00$ & $10.28$ & $22.41$ & $16.30$ & $1.20$ \\
    	\textbf{} & \textbf{$\mathrm{Pure\ SFM}$} & $54.00$ & $14.62$ & $25.45$ & $12.44$ & $\mathbf{1.25}$ \\
    	\textbf{} & \textbf{$\mathrm{SARL}$} & $62.00$ & $11.96$ & $20.69$ & $14.14$ & $1.18$ \\
    	\textbf{} & \textbf{$\mathrm{NMPC}$} & $87.00$ & $9.87$ & $\underline{16.59}$ & $12.27$ & $1.10$ \\
    	\textbf{} & \textbf{SFM-NMPC} & $\mathbf{98.00}$ & $10.35$ & $18.52$ & $\mathbf{10.70}$ & $\underline{1.24}$ \\
    	\bottomrule
    \end{tabular}
    }
\end{table}

\subsection{Quantitative Metrics Evaluation}
\begin{figure}[ht]
    \centering
    \includegraphics[width=\linewidth]{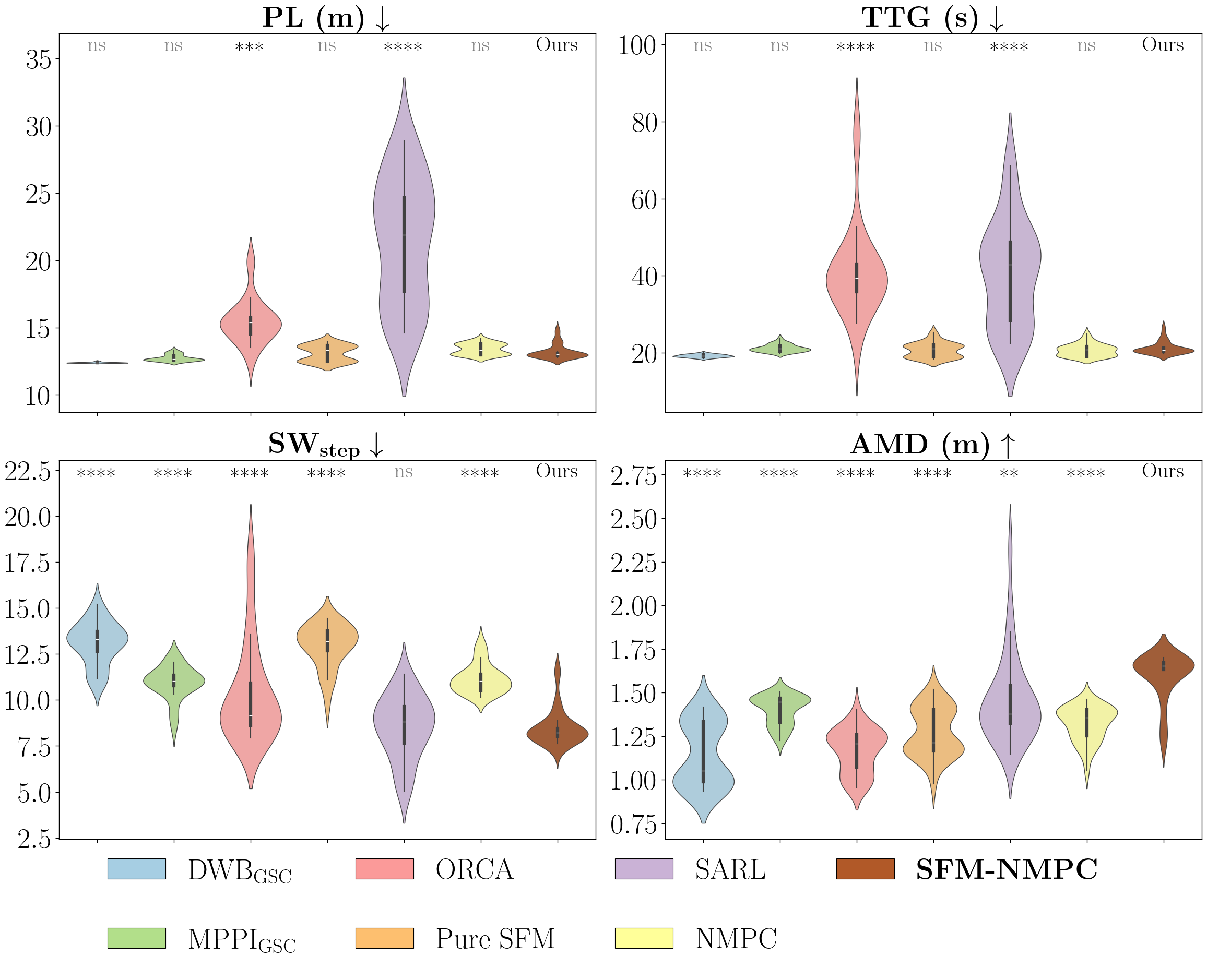}
    \caption{\footnotesize{Violin plot of the results on the open space environment. SFM-NMPC exhibits the best social behavior while maintaining navigation efficiency. (*) indicate the p-values from the pairwise Tukey HSD test.}}
    \label{fig:violin_plots}
\end{figure}

Distributions of metrics across all runs in the crowded open-space scene are compared in Fig. \ref{fig:violin_plots}.
A detailed overview of the average results is presented in Table \ref{tab:algorithm_metrics}.

\begin{figure*}[t]
  \centering
  \begin{subfigure}[t]{0.3\textwidth}
    \centering
    \includegraphics[width=\linewidth]{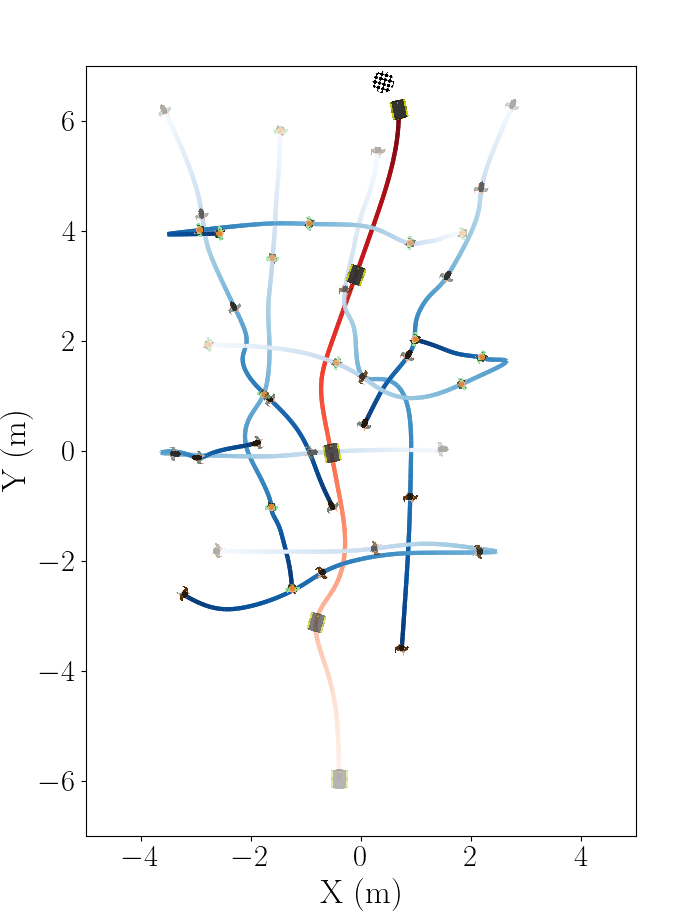}
    \caption{\textbf{SFM-NMPC} (Ours)}
  \end{subfigure}
  \hfill
  \begin{subfigure}[t]{0.3\textwidth}
    \centering
    \includegraphics[width=\linewidth]{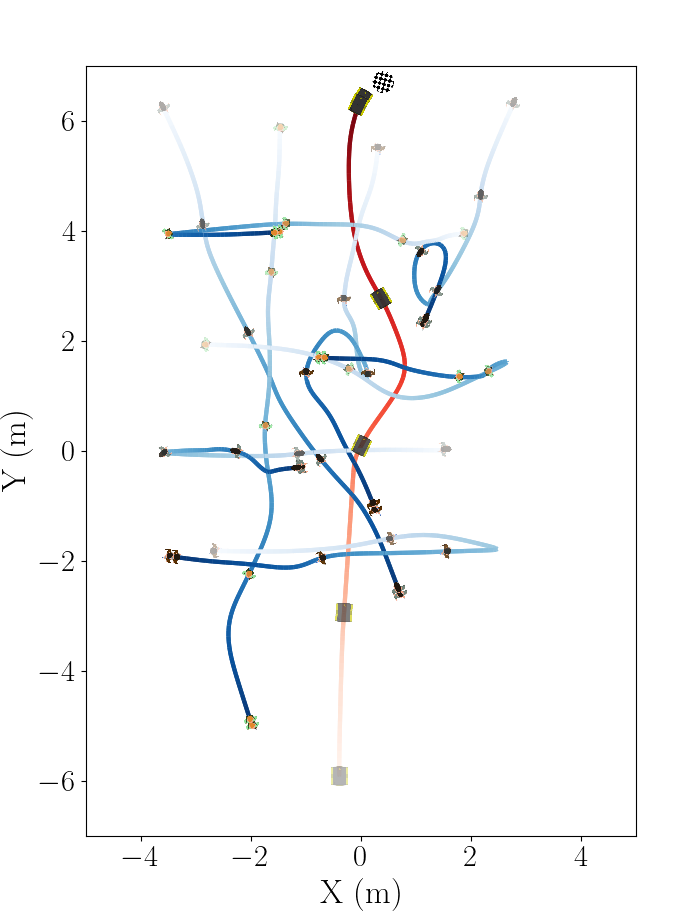}
    \caption{$\mathrm{MPPI_{GSC}}$}
  \end{subfigure}
  \hfill
  \begin{subfigure}[t]{0.3\textwidth}
    \centering
    \includegraphics[width=\linewidth]{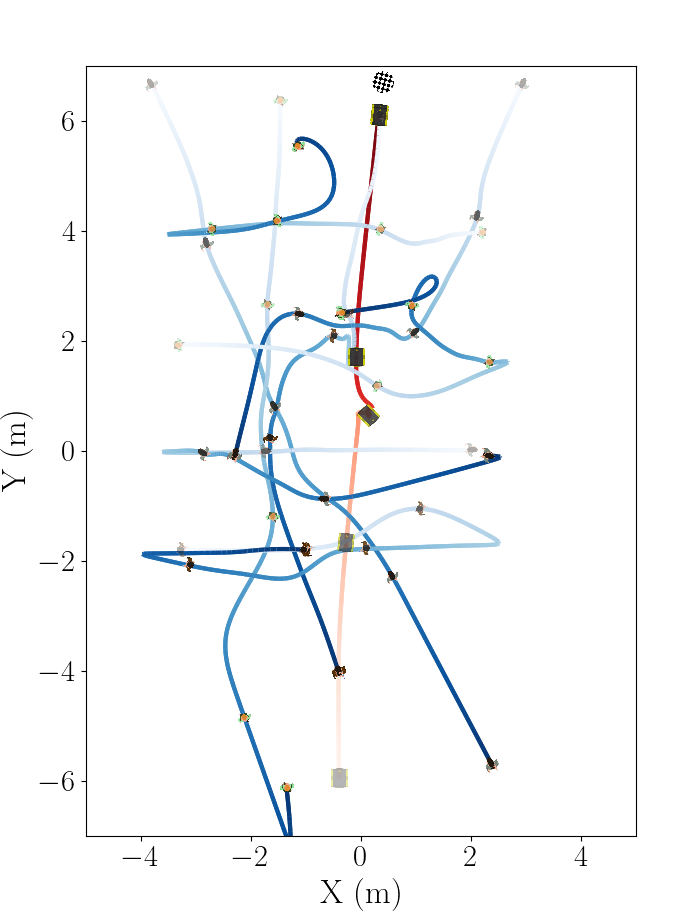}
    \caption{ORCA}
  \end{subfigure}

  \par\medskip

  \begin{subfigure}[t]{0.3\textwidth}
    \centering
    \includegraphics[width=\linewidth]{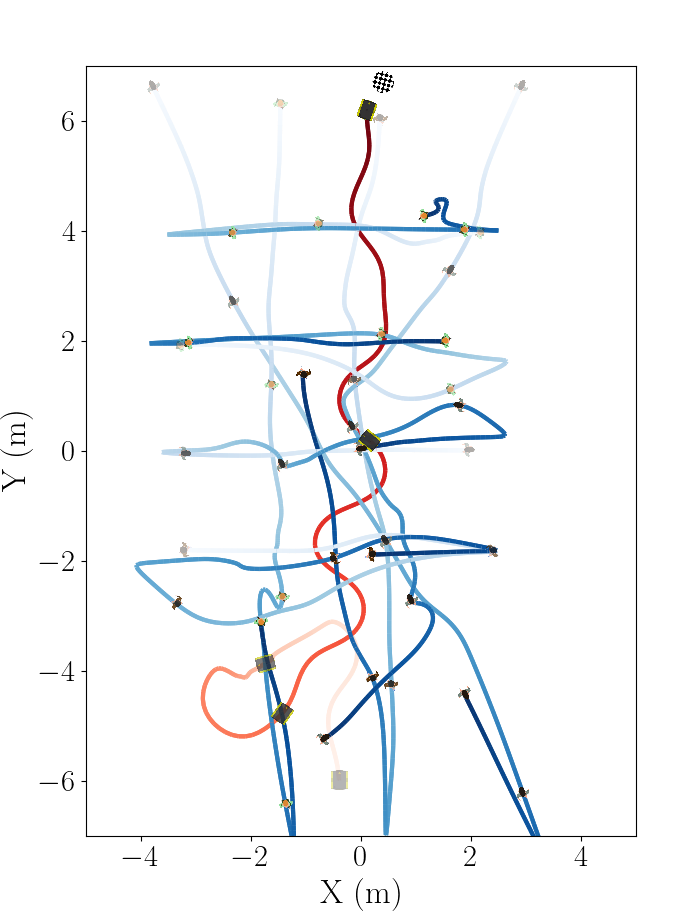}
    \caption{SARL}
  \end{subfigure}
  \hfill
  \begin{subfigure}[t]{0.3\textwidth}
    \centering
    \includegraphics[width=\linewidth]{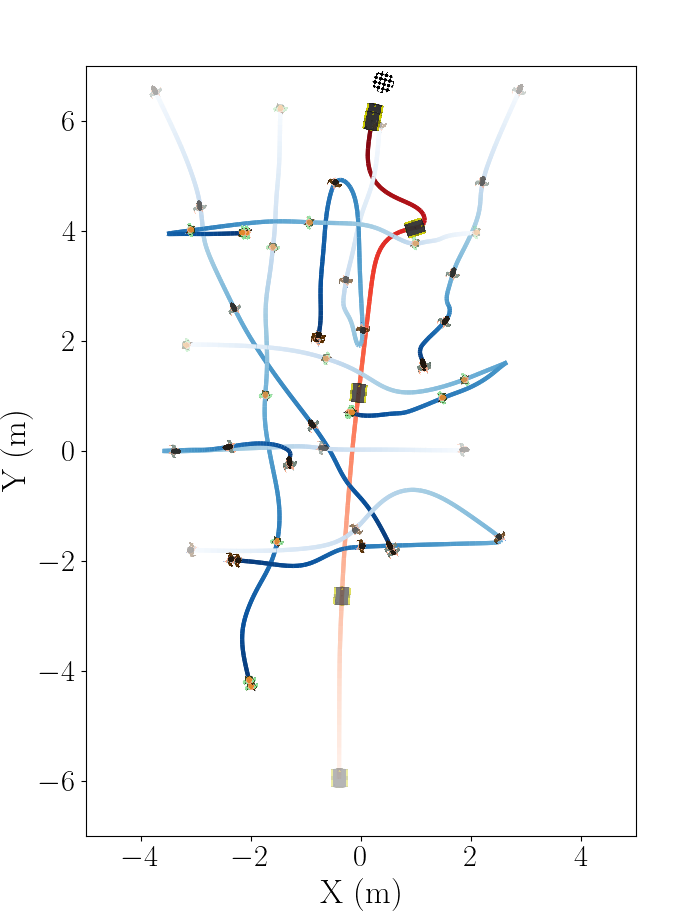}
    \caption{Pure SFM}
  \end{subfigure}
  \hfill
  \begin{subfigure}[t]{0.3\textwidth}
    \centering
    \includegraphics[width=\linewidth]{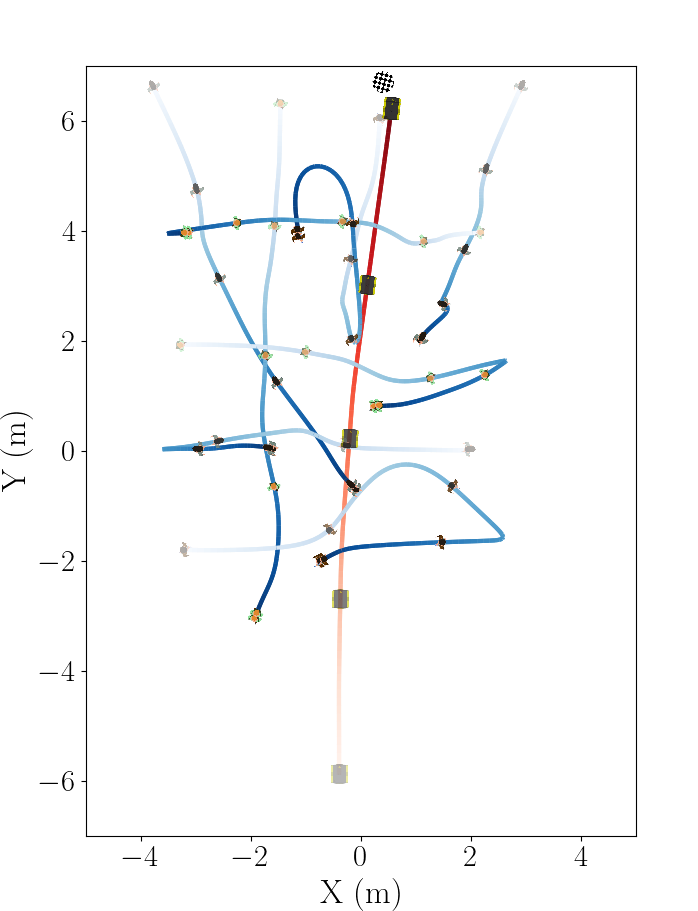}
    \caption{$\mathrm{DWB_{GSC}}$}
  \end{subfigure}
  \caption{\footnotesize{Comparison of trajectories of robot (red) and humans (blue) on the open space crowded map. Color shade indicates the temporal evolution of the experiment.}}
\label{fig:trajectories}
\end{figure*}

The average results in the open-space scenario reveal a generally high capability of all algorithms to traverse the room and reach the goal, with the worst success rate achieved by ORCA. The $\mathrm{DWB_{GSC}}$ applies the fastest navigation policy. The proposed SFM-NMPC achieves the best results in terms of the minimum distance to humans and ranks second in $SW_{step}$. However, the better score obtained by the SARL in this case must be critically evaluated, considering the path length and time-to-goal, as clearly shown in a typical trajectory in Figure \ref{fig:trajectories}d. Also, the ORCA algorithm presents almost twice the traversal time compared to the others.
In the mixed indoor scenes, corridors and narrow passages increase the difficulty. Indeed, most
baselines perform poorly, colliding with obstacles or people and eventually failing to reach the goal. The best algorithms in these challenging settings are SFM-NMPC and $\mathrm{MPPI_{GSC}}$, revealing the advantage of model predictive methods when a small reaction time to moving agents is mandatory. Also, the basic NMPC implementation without social costs performs robustly in terms of navigation success. With the integration of SFM dynamics and social soft constraints, the algorithm achieves the best $SW_{step}$, also in the overall evaluation.
Fig. \ref{fig:violin_plots} shows the distributions of the metrics as violin plots, along with pairwise comparisons from Tukey’s HSD test with our algorithm, with statistical significance denoted by asterisks corresponding to p-value thresholds following the GraphPad convention. The plots illustrate a consistent behavior across runs, with the SFM-NMPC improving over the baselines in terms of social compliance and robustness.

\begin{figure}[ht]
    \centering
    \includegraphics[width=\linewidth]{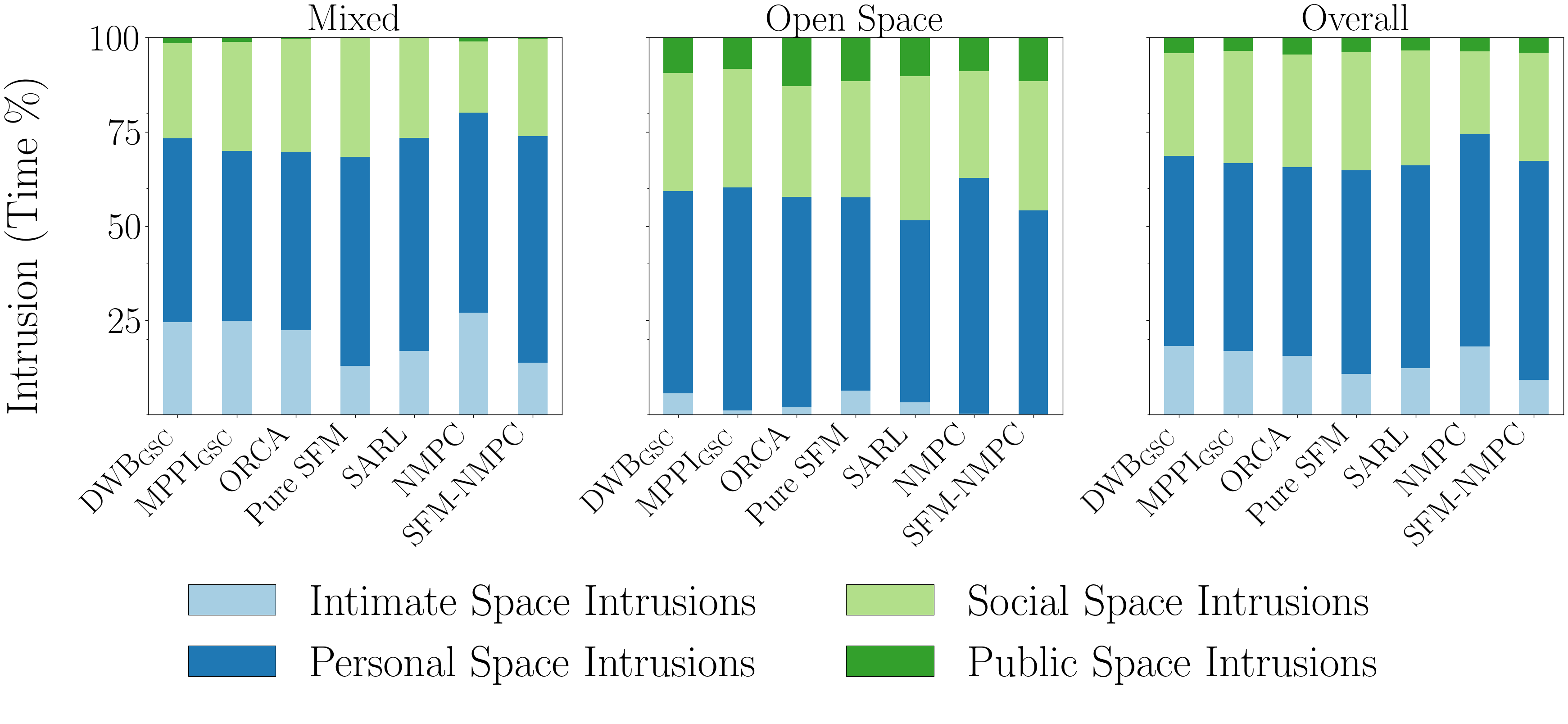}
    \caption{\footnotesize{Proxemics metrics across Mixed, Open Space, and Overall scenarios. Each bar shows the percentage of time spent intruding into different proxemic zones during the experiment.}}
    \label{fig:proxemics}
\end{figure}

\subsection{Proxemics}
Figure \ref{fig:proxemics} shows the proxemics scores: the percentage of time spent in each of Hall's proxemic zones around humans~\cite{hall1966hidden}, accumulated over the open space map, the mixed map, and overall. The SFM-NMPC demonstrates safe navigation and acceptable intrusion level, improving upon most of the compared controllers. In particular, the intrusion into the intimate space is the lowest overall, yielding performance comparable to that of the Pure SFM algorithm in the mixed indoor scenes. Thus, the proposed planner acts as the human model itself, providing a much more reliable obstacle avoidance and navigation success. In open space settings, model predictive-based planners (MPPI and NMPC) prove to be the most competitive, lowering the intimate intrusion level to nearly zero.

\begin{figure}[hb]
    \centering
    \includegraphics[width=\linewidth]{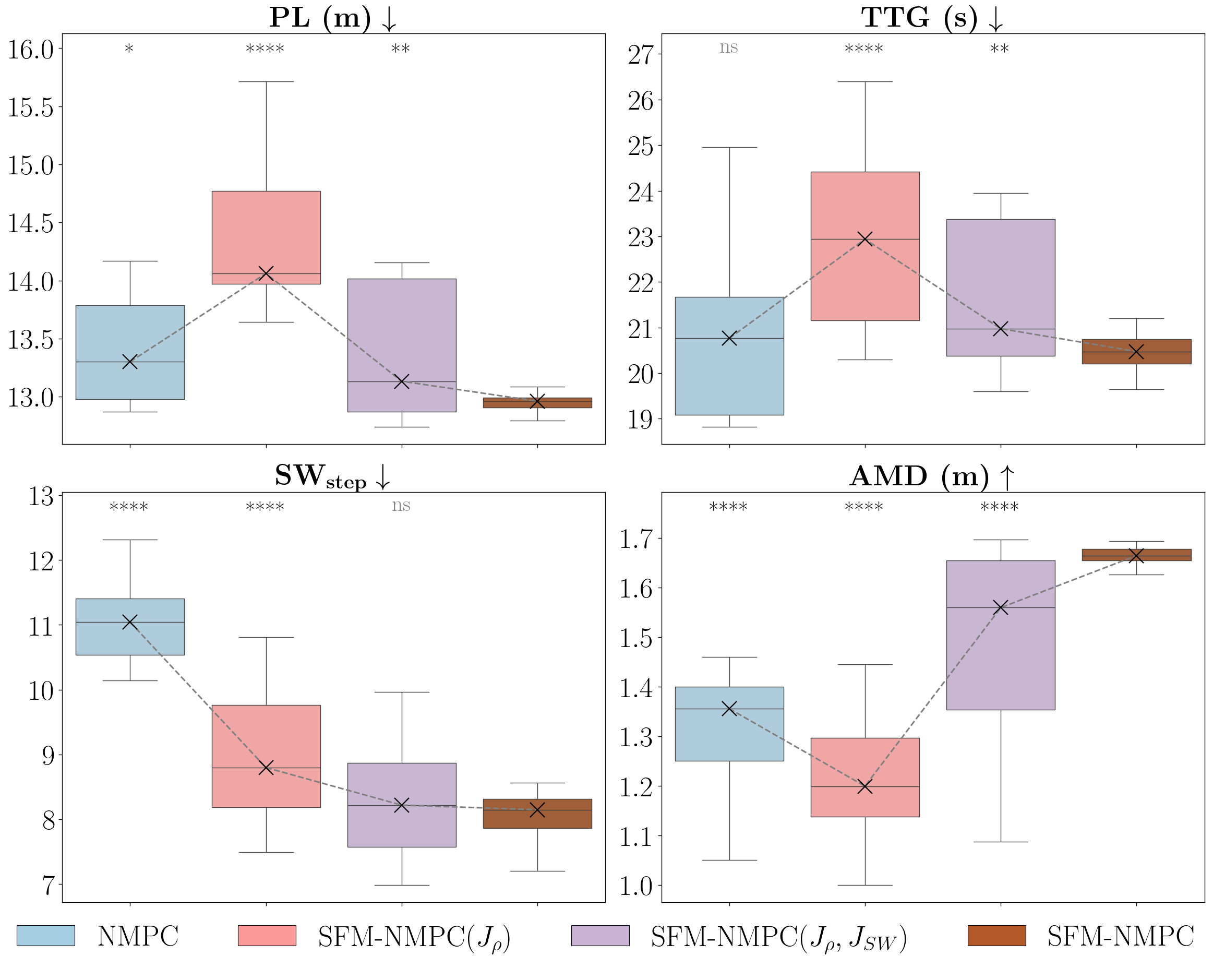}
    \caption{\footnotesize{Box plot of the results on the open space crowded environment. SFM-NMPC exhibits the best social behavior while maintaining navigation efficiency. (*) indicate the p-values from the pairwise Tukey HSD test.}}
    \label{fig:ablation_plots}
\end{figure}

\subsection{Ablation study}
An ablation study is presented to investigate the impact of the different social cost terms on the performance of the proposed SFM-NMPC.
Figure \ref{fig:ablation_plots} shows the resulting box plots, highlighting the variation in the metrics as the different social cost terms are incrementally integrated in the optimization. NMPC alone is the basic implementation without any social constraint; hence, it achieves the most efficient navigation in terms of path length and time, but the worst social impact. SFM-NMPC$(J_{\rho})$ and SFM-NMPC$(J_{\rho}, J_{SW})$ integrate proxemics and social work costs, yielding gradual improvements of both $SW_{step}$ and AMD. Nonetheless, the major boost in social metrics, measured by the median value and the robustness (as indicated by the box plot standard deviation), is provided by the addition of the heading social cost term $J_{s\theta}$. The overall SFM-NMPC, indeed, presents a significant performance improvement, also in terms of navigation efficiency.
Building on these results, we emphasize that the proposed cost function formulation constitutes a fundamental and modular component of the framework, independently contributing to performance gains regardless of the specific dynamics prediction model.

\subsection{Visual Evaluation}
Social navigation is often difficult to assess using only numerical indicators, given the intrinsic complexity of evaluating overall comfort and navigation performance. Hence, providing visual evidence of the trajectory is fundamental.
Figure \ref{fig:trajectories} shows representative trajectories obtained in the open space crowded map, visually comparing the SFM-NMPC with the baselines. The color shade indicates the temporal evolution of the motion, with the robot path in red and the human agents in blue.
First, from the trajectory analysis we can notice that the SARL algorithm, which recorded the best $SW_{step}$ result in Table \ref{tab:algorithm_metrics}, due to the high time required to reach the goal, drives the robot with an uncertain and heavily oscillating behavior typical of RL policies. From a practical perspective, a smoother and more efficient trajectory is preferred. The DWB cuts straight to the goal with limited consideration of approaching humans, resulting in the riskiest policy. ORCA and Pure SFM drive the robot straight in the first lower half of the map, tackling the person crossing only slowing down. Then, abrupt stop and inversion maneuvers are performed. The smoothest and safest trajectories are achieved by MPPI and by SFM-NMPC. Nonetheless, the only algorithm that succeeded in avoiding forcing the first person to deviate from its direction while crossing the robot's path horizontally is the SFM-NMPC, moving the robot to the left. This behavior results from the custom heading social cost adopted. Reduced interference to human motion also holds for the upper half of the map.

\section{Conclusions}
This paper presented SFM-NMPC, a non-linear Model Predictive Control framework for socially aware robot navigation that integrates the Social Force Model directly into the prediction and optimization process. By embedding human motion dynamics within the NMPC formulation, the proposed method jointly predicts the evolution of both robot and human trajectories over the planning horizon, enabling predictive decision-making in crowded environments.
Unlike approaches that rely on precomputed trajectories or purely data-driven predictors, SFM-NMPC maintains interpretability, explicitly handles constraints, and ensures real-time feasibility. The introduction of tailored social cost terms further guides the optimization toward socially compliant behaviors. Importantly, the controller achieves a \qty{20}{\Hz} execution rate despite the increased model complexity, demonstrating its suitability for real-world deployment.
Extensive experimental validation across diverse crowded scenarios confirmed the effectiveness of the proposed framework, showing improvements over competitive baselines on social compliance metrics without affecting navigation efficiency.
Future work will investigate the performance of the algorithm on a real robot, including perception uncertainty due to people and obstacles as additional noise in the optimization. Human-based metrics and evaluation protocols will be considered for the real settings. Moreover, a possible extension of the method will be investigated by exploring the co-optimization of robot and human trajectories to further refine the dynamics model.

\bibliographystyle{IEEEtran} 
\bibliography{mpc}

\end{document}